\documentclass[10pt,twocolumn,letterpaper]{article}

\usepackage{btas}
\usepackage{times}
\usepackage{epsfig}
\usepackage{graphicx}
\usepackage{amsmath}
\usepackage{amssymb}
\usepackage{multirow}
\usepackage{tabularx}
\usepackage{color}
\usepackage{subfigure}
\def\degree{${}^{\circ}$}
\graphicspath{{fig/}{cat/}{latent/}}
\graphicspath{{fig/}}

\btasfinalcopy 

\ifbtasfinal\fi
\begin{document}

\title{Latent fingerprint minutia extraction using fully convolutional network}

\author{Yao Tang \quad Fei Gao \quad Jufu Feng*\\
Key Laboratory of Machine Perception (MOE), School of EECS, Peking University\\
{\tt\small \{tangyao@, goofy@, fjf@cis.\}pku.edu.cn}
}

\maketitle
\thispagestyle{empty}

\begin{abstract}
Minutiae play a major role in fingerprint identification. Extracting reliable minutiae is difficult for latent fingerprints which are usually of poor quality. As the limitation of traditional handcrafted features, a fully convolutional network (FCN) is utilized to learn features directly from data to overcome complex background noises. Raw fingerprints are mapped to a correspondingly-sized minutia-score map with a fixed stride. And thus a large number of minutiae will be extracted through a given threshold. Then small regions centering at these minutia points
are entered into a convolutional neural network (CNN) to reclassify these minutiae and calculate their orientations. The CNN shares convolutional layers with the fully convolutional network to speed up. 0.45 second is used on average to detect one fingerprint on a GPU.
On the NIST SD27 database, we achieve 53\% recall rate and 53\% precise rate that outperform many other algorithms. Our trained model is also visualized to show that we have successfully extracted features preserving ridge information of a latent fingerprint.

\end{abstract}

\section{Introduction}

Fingerprint identification is the most historic, developed and widely used biometric technology. Everyone shares different fingerprint lines, especially minutiae which include ridge ending and ridge bifurcation.
An automatic fingerprint identification system (AFIS) is concerned with some issues including image acquisition, features extraction and matching~\cite{jain1997line}. As for latent fingerprints, which are usually obtained from crime scenes and usually have fuzzy ridges and insufficient finger regions, the traditional algorithms don't work well on them.

Generally, a fingerprint of good quality can be easily segmented and minutiae is easily extracted from it. Jain~\cite{jain1997line} follows the simple idea of ridge extraction, thinning and minutia extraction. But in latent fingerprints, fingerprint region is difficult to locate and ridge is fuzzy. So algorithms based on ridge extraction and thinning do not work well in latent fingerprints. Feng~\cite{feng2014high} proposed a minutia extraction algorithm based on Gabor feature. To reduce the influence of creases and noises, they extract minutia through Gabor phase field and measure minutiae reliability through Gabor Amplitude field. But limited by handcrafted features, the minutia descriptor still can't overcome the complex noises.

\begin{figure}[t]
 \centering
 \includegraphics[width=\linewidth]{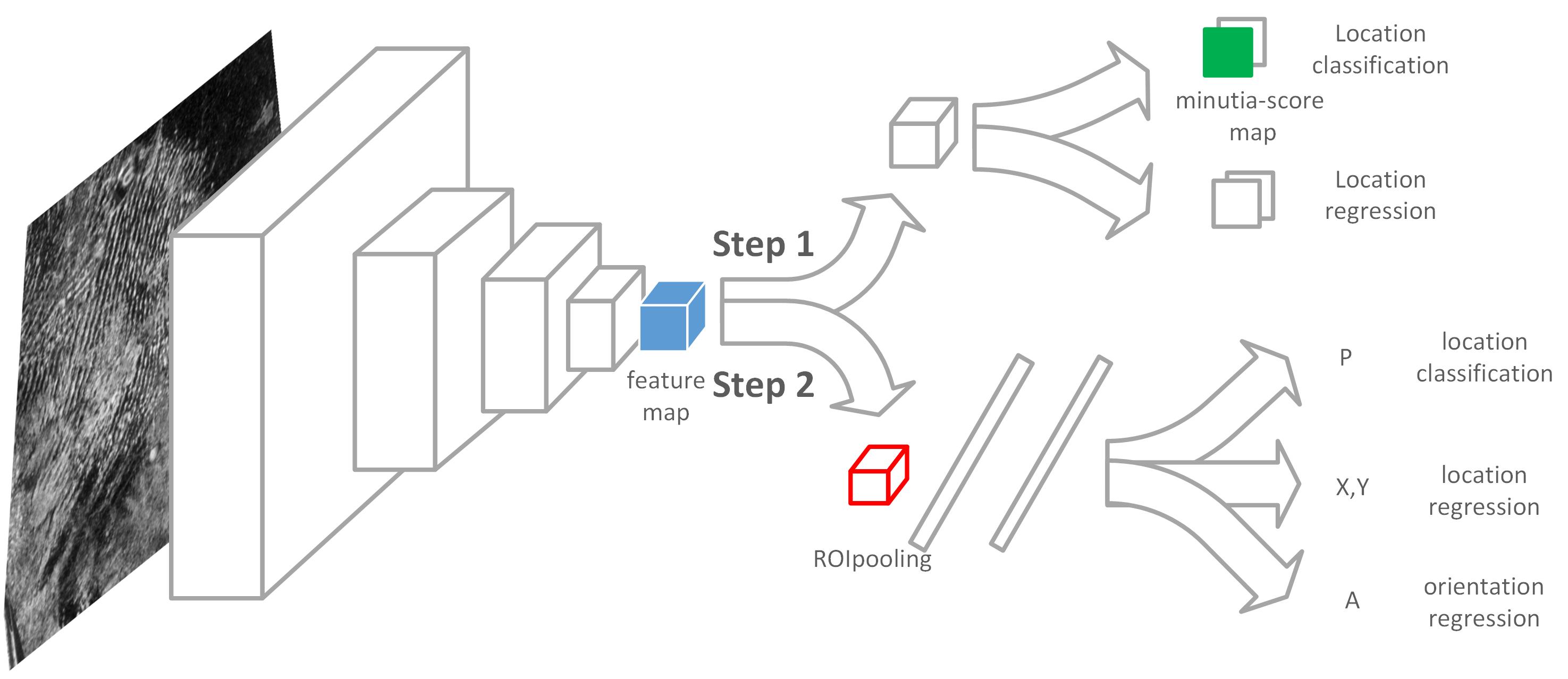}
 \caption{The architecture of our network. In step 1 FCN is used to generate proposals in pixel level, and in step 2 CNN is used to classify region-based minutiae and calculate their orientations.}
 \label{fig:overall}
\end{figure}

Considering the limitation of handcrafted features, representation learning is proved to be able to extract reliable and robust features from data.
Sankaran~\cite{sankaran2014latent} uses stacked denoising sparse AutoEncoders to extract minutiae. The minutia and non-minutia descriptors learned from a larger number of tenprint fingerprints are used to extract minutia. However they learn the descriptors from minutia patches, which lost some global information. And the orientation, which is really important in minutia matching, is ignored. Some algorithms focus on the enhancement of latent fingerprints~\cite{cao2014segmentation,yang2014localized,cao2015latent} because of the importance of orientation field in minutia extraction and even fingerprint identification.

Inspired by the success of object detection on natural images, minutia extraction is regarded as a point detection problem. Overfeat~\cite{sermanet2013overfeat} presents a framework for convolutional network to solve detection problem. Region-based convolutional network (R-CNN)~\cite{girshick2014rich} attempts to solve the problem through two steps. Firstly some category-independent proposals are generated through selective search~\cite{uijlings2013selective}. Secondly the proposals are classified using linear SVMs. In Fast R-CNN\cite{girshick2015fast}, convolutional network is used to classify the proposals, and features are extracted directly through raw image instead of every proposal. So a better results is achieved with less time cost. Considering the limitation of selective search, Faster R-CNN~\cite{ren2015faster}uses fully convolutional network~\cite{long2015fully} to generate proposals, so the net is trained end-to-end to generate proposals. The region proposal network shares full-image convolutional layers with detection network, thus the proposal network is nearly cost free.

In order to extract reliable minutiae directly from raw latent fingerprint, we propose a novel algorithm based on fully convolutional network to extract minutiae effectively and calculate their orientations. Our algorithm can be summarized as two main steps. (i) Generally proposals from raw fingerprints in pixel level using fully convolutional network. Each point at the raw fingerprint will get a minutia-like score, and a threshold is used to classify minutiae and non-minutiae. Location regression is used as a multi loss task to get more accurate locations of minutiae. (ii) Classifying the proposals and calculating their orientations. The orientation is calculated through a multi-task loss layer consisting of minutia classification, location regression and orientation regression.

The key contributions of this paper are as follows:
\begin{enumerate}
    \item Minutiae extraction is seen as a point detection problem and the features are learned from data automatically to adapt to complex background.
    \item Minutiae descriptor is learned end-to-end that no middle process is required.
    \item Reliable minutiae are extracted directly from raw fingerprints without segmentation or enhancement.
\end{enumerate}

\section{Proposed minutia extraction algorithm}
Minutia is a special structure in fingerprint including ridge ending and bifurcation. Fully convolutional network (FCN) is used to map raw fingerprints to a minutia-score map with a fixed stride. Minutia-score map generates proposals in pixel level through a given threshold. Then a CNN is learned to classify these proposals and calculate their orientations. The architecture of our network is shown in Fig.~\ref{fig:overall}. Our algorithm follows two steps, (i) Generating proposals  in pixel level using FCN. (ii) Classifying the regions centering at proposed minutiae and calculating their orientations using CNN.

\subsection{Generating proposals with fully convolutional network}

FCN is used to map raw fingerprint to minutia-score map with a fixed stride. Minutia-score map is the output of the network, and each point is a minutia-like score corresponding to the raw fingerprints.
In this paper, we set fixed stride as 16. It is considered as the tolerance scope of location distance. As shown in Fig.~\ref{fig:overall}, FCN is trained to generate proposals in raw fingerprints in step 1. Every 16*16 pixel region (under 500 pixels per inch) in raw fingerprints is mapped to one single point in minutia-score map. The score, ranging from 0 to 1, gets bigger when it's more likely to contain minutiae in the corresponding region.
Thus a threshold can be used to generate proposals. Location regression is also used as a multi-task loss to get more accurate locations of minutiae.

\subsubsection{FCN architecture}

The architecture we used is formed of two parts. The former convolutional layers are showed in Table.~\ref{table:SHAREDCONVLAYERS}. In our practice, three pretrained models are used to initialize the former convolutional layers, including ZF model~\cite{zeiler2014visualizing}, VGG model~\cite{simonyan2014very} and deep residual model~\cite{he2015deep}.
Remaining layers consists of a classification layer and a location regression layer, and they are showed in Table.~\ref{table:RPN_FASTRCNN}.

\setlength{\tabcolsep}{4pt}
\begin{table}[t]
\begin{center}

\scriptsize
\scalebox{0.76}{
\begin{tabular}{cccc}
\hline\noalign{\smallskip}
input & \multicolumn{3}{c}{images with maxsize (n, m)} \\
\noalign{\smallskip}\hline\noalign{\smallskip}
size & ZF model & VGG moel & 50-residual model\\
\noalign{\smallskip}
\hline
\noalign{\smallskip}
$1$
&
& $\left[\begin{tabular}[t]{c}$3\times 3$ conv, 64\end{tabular}\right]\times 2$
\\
\noalign{\smallskip}
\hline
\noalign{\smallskip}
$1/2$
& \begin{tabular}{c}$7\times 7$ conv, 96, /2\end{tabular}
& \begin{tabular}{c}$2\times 2$ max pool, /2 \\ $\left[\begin{tabular}{c}$3\times 3$ conv, 128\end{tabular}\right]\times 3$ \end{tabular}
& \begin{tabular}{c}$7\times 7$ conv, 64, /2\end{tabular}
\\
\noalign{\smallskip}
\hline
\noalign{\smallskip}
$1/4$
& \begin{tabular}{c}$3\times 3$ max pool, /2  \end{tabular}
& \begin{tabular}{c}$2\times 2$ max pool, /2 \\ $\left[\begin{tabular}{c}$3\times 3$ conv, 256\end{tabular}\right]\times 3$ \end{tabular}
& \begin{tabular}{c} $\left[\begin{tabular}{c}$1\times 1$ conv, 64 \\ $3\times 3$ conv, 64 \\ $1\times 1$ conv, 256\\\end{tabular}\right]\times 3 $\end{tabular}
\\
\noalign{\smallskip}
\hline
\noalign{\smallskip}
$1/8$
& \begin{tabular}{c}$5\times 5$ conv, 256, /2\end{tabular}
& \begin{tabular}{c}$2\times 2$ max pool, /2 \\ $\left[\begin{tabular}{c}$3\times 3$ conv, 512\end{tabular}\right]\times 3$ \end{tabular}
& \begin{tabular}{c} $\left[\begin{tabular}{c}$1\times 1$ conv, 128 \\ $3\times 3$ conv, 128 \\ $1\times 1$ conv, 512\\\end{tabular}\right]\times 4 $\end{tabular}
\\
\noalign{\smallskip}
\hline
\noalign{\smallskip}
$1/16$
& \begin{tabular}{c}$3\times 3$ max pool, /2
	\\ $\left[\begin{tabular}[t]{c}$3\times 3$ conv, 384\end{tabular}\right]\times 2$
	\\ $\left[\begin{tabular}[t]{c}$3\times 3$ conv, 256\end{tabular}\right]\times 1$\end{tabular}
& \begin{tabular}{c}$2\times 2$ max pool, /2 \\ $\left[\begin{tabular}{c}$3\times 3$ conv, 512\end{tabular}\right]\times 3$ \end{tabular}
& \begin{tabular}{c} $\left[\begin{tabular}{c}$1\times 1$ conv, 256 \\ $3\times 3$ conv, 256 \\ $1\times 1$ conv, 1024\\\end{tabular}\right]\times 6 $\end{tabular}
\\
\noalign{\smallskip}
\hline
\noalign{\smallskip}
output &
\multicolumn{3}{c}{features with maxsize (n/16, m/16)} \\
\noalign{\smallskip}
\hline

\end{tabular}
}
\setlength{\tabcolsep}{1.4pt}
\setlength{\belowcaptionskip}{10pt}
\caption{
	Architectures of shared convolutional layers. Column `size' refers to downsampling scales of input images. For residual-net frameworks, downsampling is performed by the first conv layer at each scale, with stride of 2.
}
\label{table:SHAREDCONVLAYERS}
\end{center}
\end{table}

\subsubsection{Loss function}

Considering the FCN maps raw fingerprints to minutia maps, rectangular regions are mapped into a single point (eg. 16*16 to 1). Thus a single point in minutia map corresponds to a rectangular region. To get more accurate locations, location regression is used to calculate the location deviation. The loss function is a multi-task loss as below:
\begin{equation}
    L(p,p^{*},t,t^{*}) = L_{cls}(p,p^{*})+\lambda p^{*} L_{loc}(t,t^{*})
\end{equation}

Here $p$ and $t$ represents the predicted probability and location coordinates. The symbol with star means its corresponding ground truth.
Let $p^{*}$ equals to 1 if the corresponding region contains manually marked minutiae.
$p^{*}$ multiplied to $L_{loc}$ means that only minutia point is concerned to contribute to the loss. The classification loss $L_{cls}$ is log loss over two classes. $L_{loc}$ is defined in ~\cite{girshick2015fast} as a smooth L1 loss. It is showed in eq.~\ref{eq:L2}.
\begin{equation}\label{eq:L2}
    L_{loc}(t,t^{*})=
    \begin{cases}
        0.5(t-t^{*})^2 \quad if |t-t^{*}|<1 \\
        |t-t^{*}|-0.5  \quad otherwise
    \end{cases}
\end{equation}

\subsubsection{Training}

\begin{figure}[t]
 \centering
 \includegraphics[width=\linewidth]{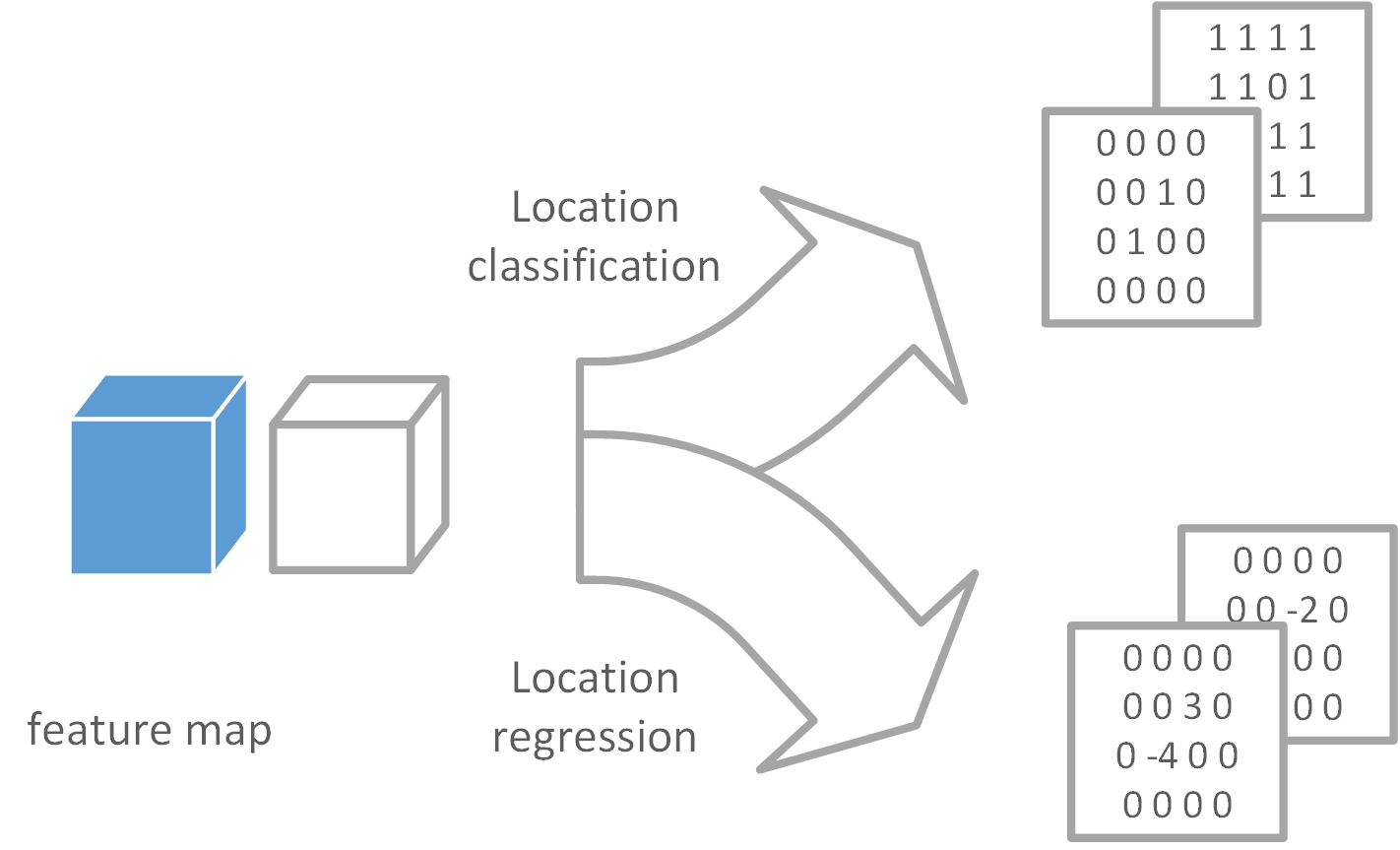}
 \caption{A toy sample of FCN training. There are two minutiae point in feature map with coordinates $(2,3)$ and $(3,2)$, and their distance to ground truth minutia is $(3,-2)$ and $(-4,3)$.}
 \label{fig:trainrpn}
\end{figure}

An extracted minutia is assigned to be true if its distance to a manually labeled minutia is less than 15 pixels, and the angle between the two is less than $30^{\circ}$. We assign the center of the region as the initial value of t. Location deviation is also assigned to regress the minutia location in training to make predicted coordinates more accurate. Fig.~\ref{fig:trainrpn} shows a toy sample. The global learning rate is 0.001 and reduces to 0.0001 after 30000 mini-batch iterations. The momentum is 0.9 and parameter decay is 0.0005.
Our implementation uses Caffe~\cite{jia2014caffe}.

\setlength{\tabcolsep}{4pt}
\begin{table}[t]
\begin{center}

\scriptsize
\scalebox{0.95}{
\begin{tabular}{ccccc}
\hline
\noalign{\smallskip}
input
& \multicolumn{4}{c}{features extracted by shared conv layers} \\

\noalign{\smallskip}
\hline
\noalign{\smallskip}
type
& \multicolumn{2}{c}{FCN} & \multicolumn{2}{c}{CNN}\\

\noalign{\smallskip}
\hline
\noalign{\smallskip}

&\multicolumn{2}{c}{$\left[\begin{tabular}{c}$3\times 3$ conv, 256\end{tabular}\right]\times 1$}
& \multicolumn{2}{c}{\begin{tabular}{c}$6\times 6$ roi pooling~\cite{girshick2015fast} \\ $\left[\begin{tabular}{c}fc 4096\end{tabular}\right]\times 2$ \end{tabular}}
\\

\noalign{\smallskip}
\hline
\noalign{\smallskip}
&\begin{tabular}{c}$1\times 1$ conv, 2\end{tabular}
& \begin{tabular}{c}$1\times 1$ conv, 2\end{tabular}
& softmax 2 & fc 3
\\

\noalign{\smallskip}
\hline
\noalign{\smallskip}
output
& classification & location regression
& classification & location regression
\\

\noalign{\smallskip}
\hline
\end{tabular}
}
\setlength{\tabcolsep}{1.4pt}
\setlength{\belowcaptionskip}{10pt}
\caption{
	Remaining different layers of our network. Left side are layers cascaded to the FCN and right side are layers cascaded to the CNN. Roi pooling~\cite{girshick2015fast} is used to extracted regions of interest in feature maps.
}
\label{table:RPN_FASTRCNN}
\end{center}
\end{table}

\subsection{Region-based classification and orientation regression}

\begin{figure}[t]
 \centering
 \includegraphics[width=\linewidth]{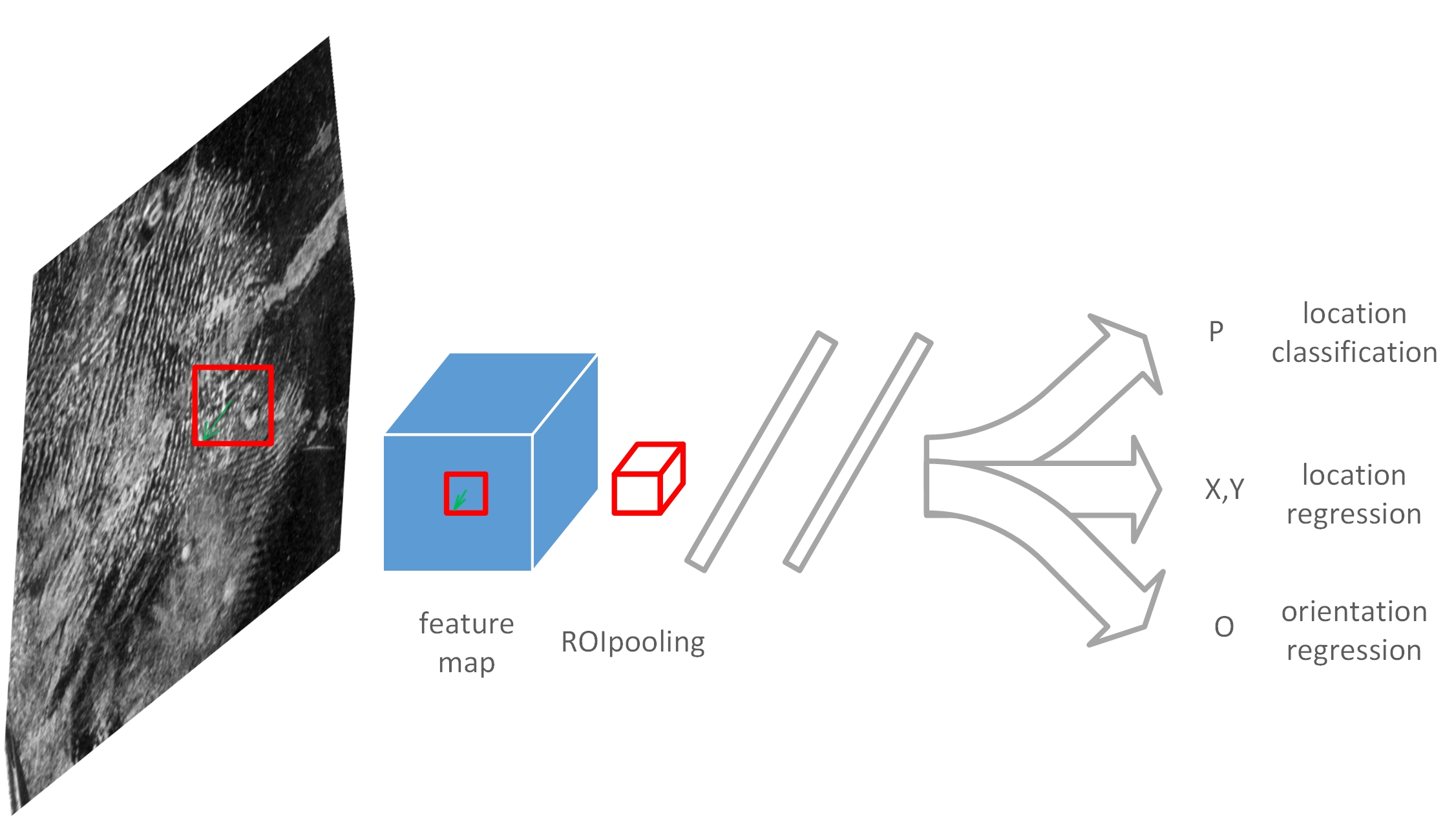}
 \caption{Flowchart of region-based classification and orientation regression. The ROI pooling layer~\cite{girshick2015fast} is used to extract corresponding regions in feature map. }
 \label{fig:fastrcnn}
\end{figure}

FCN has generated a large number of proposals without orientation, and step 2 will reclassify these proposals and calculate their orientations based on small regions. The expanded regions centering at proposal points are put into a convolutional neural network (CNN) with multi-task loss layer consisting of minutia classification, location regression and orientation regression. The proposals are expanded to 32*32 pixel regions that we consider having enough information to determine whether the proposals are minutia points.
In order to accelerate the minutiae extracting speed, we force step 1 and step 2 to use the same features extracted from raw fingerprints by sharing the same convolutional layers in Fig.~\ref{fig:overall}.
And we add some extra layers to meet different targets, as shown in Table.~\ref{table:RPN_FASTRCNN}.
Fig.~\ref{fig:fastrcnn} shows the flowchart of region-based classification and orientation regression. The minutia regions are extracted in feature maps corresponding to fingerprints in Fig.~\ref{fig:fastrcnn}.

\subsubsection{Loss function}

The loss function consists of three parts: classification, location regression and orientation regression.
It is defined as below:
\begin{equation}
\begin{aligned}
    L(p,p^{*},t,t^{*},o,o^{*}) &= L_{cls}(p,p^{*})+\lambda p^{*} L_{loc}(t,t^{*})+\\
    &\beta p^{*}L_{ori}(o,o^{*})
\end{aligned}
\end{equation}

Here $p, t,$ and $o$ represents the predicted probability, location and orientation respectively. The symbol with star means its corresponding ground truth. $p^{*}$ multiplied to $L_{cls}$ and $L_{ori}$ means that only minutia point is concerned to contribute to the loss. $\lambda$ and $\beta$ are set to be 1 and 3 in our experiments.
The classification loss $L_{cls}$ is log loss over two classes. $L_{loc}$ and $L_{ori}$ are defined in eq.~\ref{eq:L2}.

\subsubsection{Co-training}
In order to accelerate the minutiae extracting speed, FCN and CNN will share the same convolutional layers.
Inspired by the Faster R-CNN~\cite{ren2015faster}, a 4-step training is used:
\begin{enumerate}
    \item Training FCN with raw latent fingerprints and the net is pretrained by ZF net. Then proposals are generated with the FCN.
    \item Training CNN with proposals generated in step 1 and the net is pretrained by ZF net.
    \item Training a new FCN with raw latent fingerprints and the net is pretrained by CNN in step 1. And the convolutional layers are fixed.
    \item Training a new CNN with proposals generated in step 3 and the net is pretrained by CNN in step 1 too. And the convolutional layers are also fixed.
\end{enumerate}

Now the two networks shares the same convolutional layers. Then a latent fingerprint can be extracted following several simple steps: calculating feature maps, generating proposals, region-based classifying and calculating orientations.

\subsection{Non-maximum suppression}

Since our proposed algorithm detect minutia points directly from the raw fingerprints, several minutiae around a ground truth minutia will all be assigned true. Actually only one minutia will be set to be true corresponding to a manually marked minutia. To solve this problem, non-maximum suppression (NMS) is used to suppress minutiae which are too similar to a high-score minutia. In this paper, 16 pixels and 30\degree is used to clip the redundant minutia points.

NMS is used to clip the proposals generated in step 1 to assure that step 2's training data will not be clustered to several high score proposals. In our experiments this is really important for training. NMS is also used in final output to clip the redundant minutiae which matched to the same manually marked minutia.

\section{Experiments}
We compare our proposed algorithm with other algorithms in terms of minutiae extraction performance. Our trained model is also visualized to see
that we have successfully extracted features preserving ridges and valleys of a latent fingerprint.

\subsection{Database}
Large number of training examples are required to learn a deep convolution network. Therefore, an expanded latent fingerprint database was collected for training.
The database contains gray scale fingerprint images collected by China's police department from crime scenes. In all there are 4205 latent cases and corresponding minutiae data validated by professional team of latent examiners. Each image is $512\times{512}$ pixels in size and 500 pixels per inch (ppi).

NIST SD27~\cite{garris2000nist} is used to compare with other algorithms. The database has 258 latent fingerprints including annotated minutiae by experts and their matching tenprint images. Each image is $800\times{768}$ pixels in size and the ppi is also 500. Examiners subjectively assigned an overall quality, good, bad and ugly, to every latent image. The database is randomly sampled 50\% (129 images) according to their quality for training and testing.

When using both database as training set, we pad zeros down right to our expanded database in order not to change the ppi. And neither enhancement nor segmentation is carried out on latent fingerprints.

\begin{figure*}[t]
\centering     
\subfigure[raw fingerprints]{\includegraphics[width=0.3\linewidth]{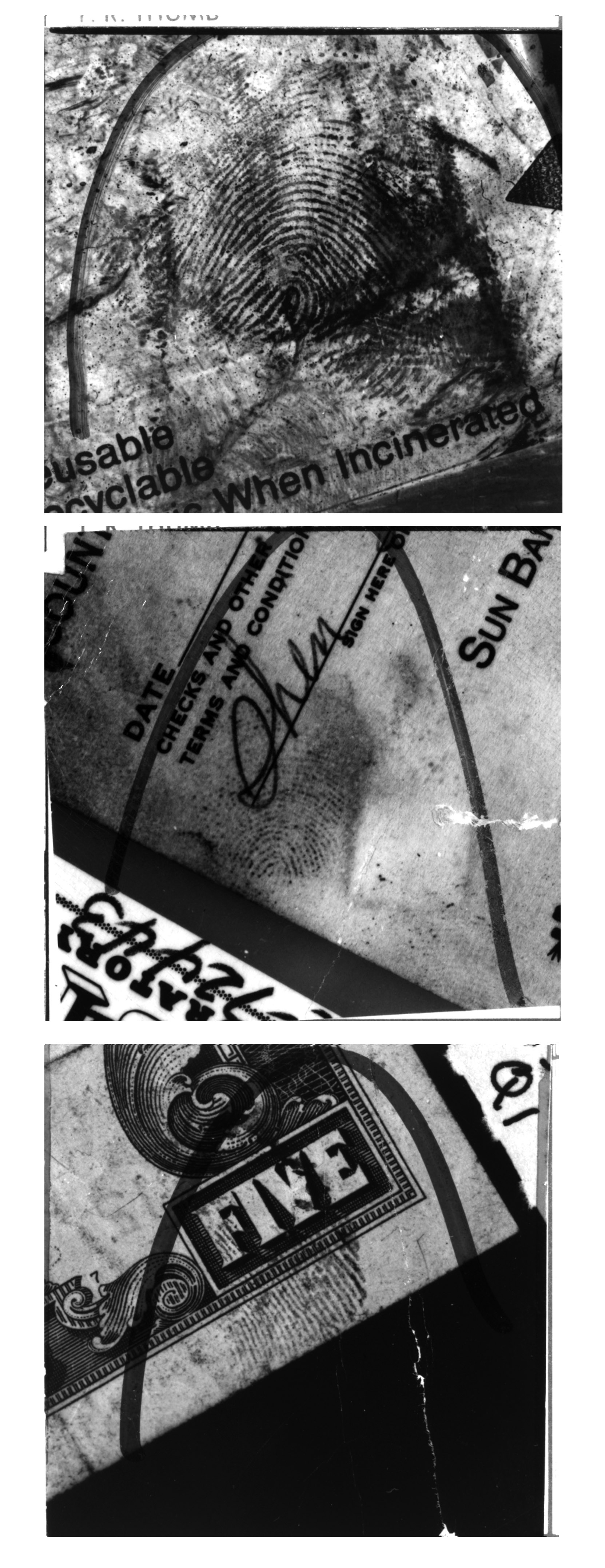}}
\subfigure[proposals]{\includegraphics[width=0.3\linewidth]{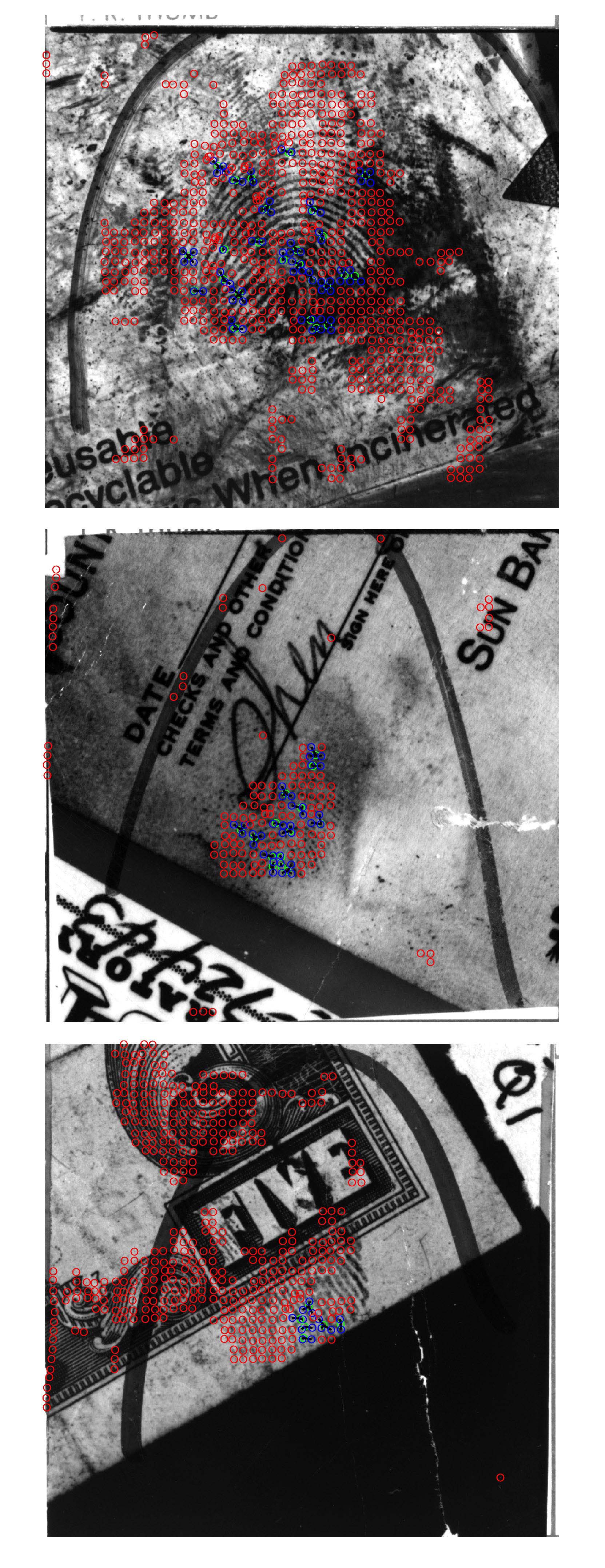}}
\subfigure[final minutiae]{\includegraphics[width=0.3\linewidth]{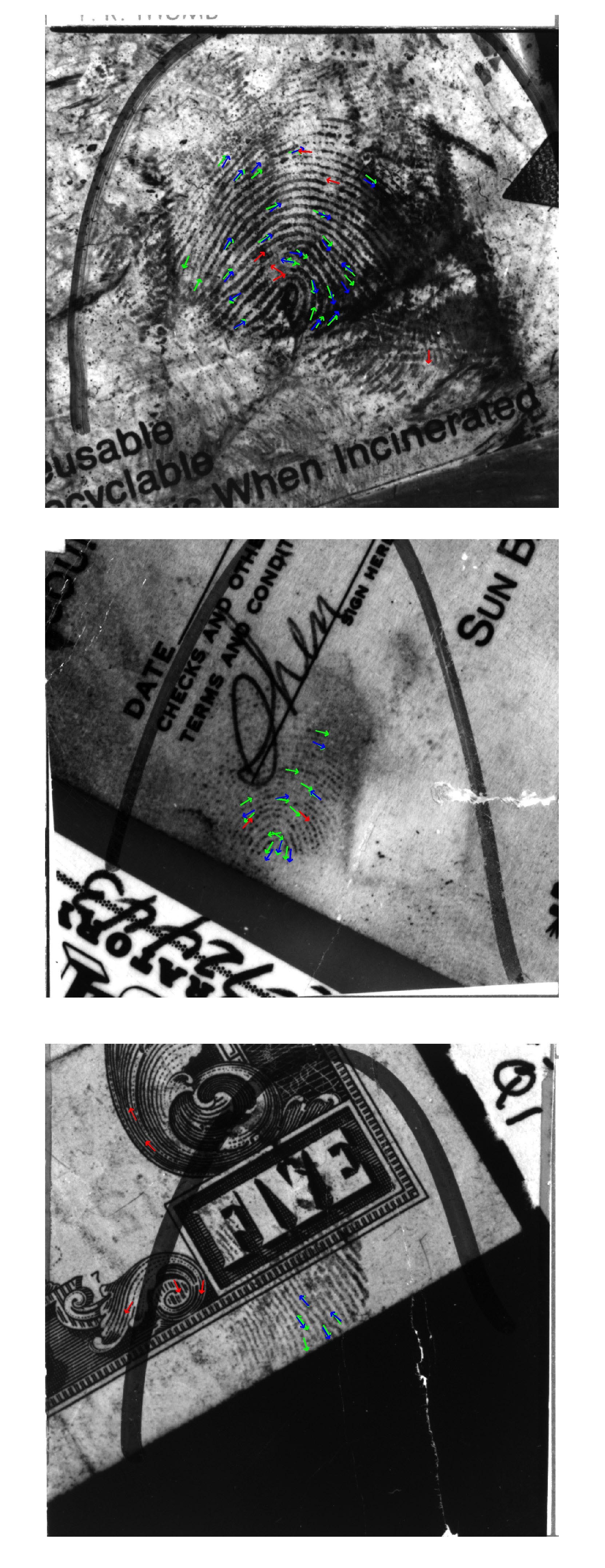}}
\caption{Minutia extraction results of our algorithm on different quality latent fingerprints in NIST SD27. It is good, bad, and ugly fingerprints from up to down. Column (a) are raw fingerprints, column (b) are proposals generated by FCN, and column (c) are final minutia points. Green, blue and red symbols stands for manually marked minutiae, correctly extracted minutiae and wrong minutiae respectively.}
\label{fig:vis_minutia}
\end{figure*}

\setlength{\tabcolsep}{4pt}
\begin{table}[t]
\begin{center}

\scriptsize
\scalebox{1.13}{
\begin{tabular}{ccccc}
\hline
\noalign{\smallskip}
Model & Recall (R) & Precision (P) & F1 score & Time cost per img\\
\noalign{\smallskip}
\hline
\noalign{\smallskip}
ZF~\cite{zeiler2014visualizing} & 36.7\% & 36.5\% & 0.3660 & 0.13s\\
\noalign{\smallskip}
\hline
\noalign{\smallskip}
VGG~\cite{simonyan2014very} & 53.0\% & 53.4\% & 0.5320 & 0.44s\\
\noalign{\smallskip}
\hline
\noalign{\smallskip}
DR50~\cite{he2015deep} & 48.7\% & 48.1\% & 0.4840 & 1.90s \\
\noalign{\smallskip}
\hline

\end{tabular}
}
\setlength{\tabcolsep}{1.4pt}
\setlength{\belowcaptionskip}{10pt}
\caption{
	Minutiae extraction performance and time cost on different pre-trained models.
}
\label{table:models_performance}
\end{center}
\end{table}

\setlength{\tabcolsep}{4pt}
\begin{table}[t]
\begin{center}

\scriptsize
\scalebox{1.25}{
\begin{tabular}{ccccc}
\hline
\noalign{\smallskip}
Algorithm & Recall & Precision & F1 score & time (s)\\
\noalign{\smallskip}
\hline
\noalign{\smallskip}
MINDTCT~\cite{watson2007user} & 18.6\% & 1.1\% & 0.0208 & 0.32\\
\noalign{\smallskip}
\hline
\noalign{\smallskip}
Gabor~\cite{feng2014high} & 7.6\% & 23.1\% & 0.1144 & 1.17\\
\noalign{\smallskip}
\hline
\noalign{\smallskip}
Gabor\&CNN & 35.1\% & 56.3\% & 0.4324 & 1.37\\
\noalign{\smallskip}
\hline
\noalign{\smallskip}
AutoEncoder~\cite{sankaran2014latent} & 65\% & 26\% & 0.3714 & -\\
\noalign{\smallskip}
\hline
\noalign{\smallskip}
Proposed algorithm & 53.0\% & 53.4\% & 0.5320 & 0.45\\
\noalign{\smallskip}
\hline

\end{tabular}
}
\setlength{\tabcolsep}{1.4pt}
\setlength{\belowcaptionskip}{10pt}
\caption{
	Minutiae extraction performance and time cost on different algorithms.
}
\label{table:m_performance}
\end{center}
\end{table}

\subsection{Minutiae extraction performance}

The performance of minutiae extraction is evaluated in terms of the true positive rate (recall), positive predictive value (precision) and F1 score~\cite{rijsbergencj}, with the manually extracted minutiae as ground truth. Typically, an extracted minutia is assigned to be true if its distance to a manually labeled minutia is less than 15 pixels, and the angle between the two is less than 30\degree. Furthermore, this is one to one match.

Table.~\ref{table:models_performance} shows the time-cost and extraction performance on different pre-trained models. DR50 model is deeper but does not get a best result because of the lack of training data and over fitting is more serious.
VGG model gets a highest accuracy and medium time cost, so it is chosen for further identification and visualization.

Table.~\ref{table:m_performance} shows the minutiae extraction performance on NIST SD27 compared to some other algorithms. MINDTCT~\cite{watson2007user} and Gabor~\cite{feng2014high} are minutia extraction algorithms on tenprints.
Gabor\&CNN generates proposals through a Gabor-based algorithm~\cite{feng2014high} with a very low threshold, typically 70.6\% of recall and 1.6\% of precision is acquired. Then a CNN is learned to classify these proposals. An AutoEncoder-based algorithm~\cite{sankaran2014latent} extracts minutiae with a learned stacked denoising spare autoencoder. The latent fingerprints are manually segmented and their orientations are not calculated. Our proposed algorithm extracts minutiae directly from raw fingerprint without segmentation or enhancement. The FCN gets a better result with about 96.6\% of recall and 12.4\% of precision compared to Gabor-based proposal algorithm.
The mean error of location and angle are 6.73 pixels and 7.7\degree respectively.
Table.~\ref{table:gbu_performance} shows the extraction performance on different quality latent fingerprints, and Fig.~\ref{fig:vis_minutia} visualize the extraction results.

\setlength{\tabcolsep}{4pt}
\begin{table}[t]
\begin{center}

\scriptsize
\scalebox{1.1}{
\begin{tabular}{cccc}
\hline
\noalign{\smallskip}
Quality & Recall (R) & Precision (P) & F1 score \\
\noalign{\smallskip}
\hline
\noalign{\smallskip}
good & 60.5\% & 61.7\% & 0.6116 \\
\noalign{\smallskip}
\hline
\noalign{\smallskip}
bad & 42.4\% & 43.4\% & 0.4292 \\
\noalign{\smallskip}
\hline
\noalign{\smallskip}
ugly & 44.0\% & 46.1\% & 0.4505 \\
\noalign{\smallskip}
\hline

\end{tabular}
}
\setlength{\tabcolsep}{1.4pt}
\setlength{\belowcaptionskip}{10pt}
\caption{
	Minutiae extraction performance on different quality latent fingerprints in NIST SD27.
}
\label{table:gbu_performance}
\end{center}
\end{table}

To investigate the influence of location regression and orientation regression in minutiae extraction, some ablation experiments are performed. Firstly, location regression in accurate classification is removed, which indicates that final location is without any fine tuning. Recall and precision are decreased to 50.0\% and 49.7\%. Secondly, only orientation regression is removed, which means that orientation is no longer a criterion to filter the minutia. Recall and precision are increased to 55.0\% and 55.4\%, which indicates that we lost few minutiae in calculating orientations.

\subsection{Identification performance}
In addition to the mated tenprints of 258 images from NIST SD27, we also include 2000 tenprint gallery from NIST SD4 database~\cite{watson1992nist}. Minutiae in NIST SD4 is extracted by extractor mindtct from NBIS~\cite{watson2007user}.
Matcher bozorth3, also from NBIS, is adopted in the identification experiments. Only minutiae information is used in this matching algorithm.

\begin{figure}[t]
 \centering
 \includegraphics[width=\linewidth]{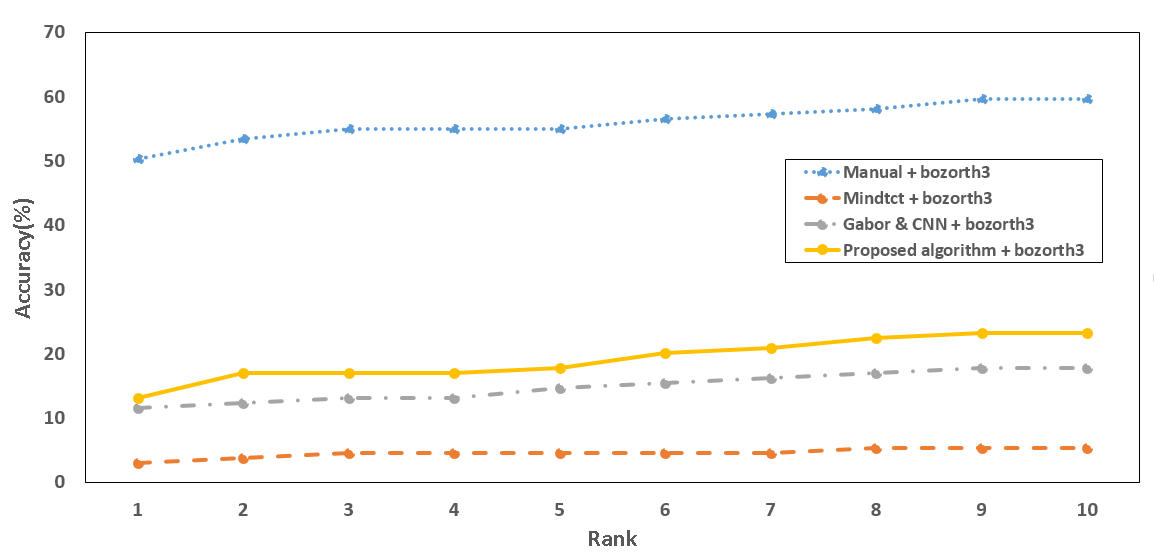}
 \caption{Identification performance of different algorithms on 129 latents from NIST SD27.}
 \label{fig:matching}
\end{figure}

As shown in Fig~\ref{fig:matching}, we compared the identification performance with minutiae extracted by mindtct, Gabor\&CNN and manually marked.
The result shows that our proposed algorithm get a better performance.

\subsection{Visualization}

\begin{figure}[t]
\centering     
\subfigure[raw fingerprints]
{\label{fig:VisualRes_a}\includegraphics[width=0.4\linewidth]{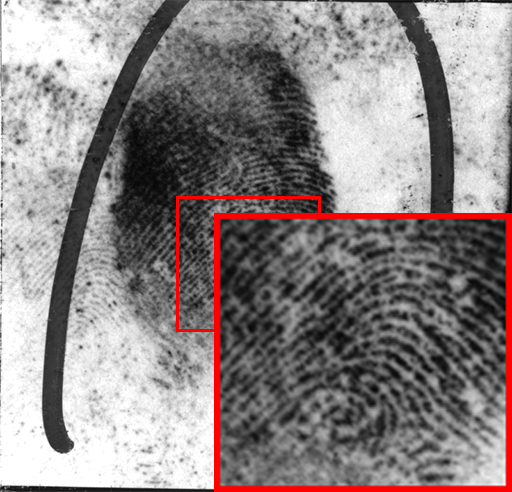}}
\subfigure[VGG conv4-3]{\label{fig:VisualRes_b}\includegraphics[width=0.4\linewidth]{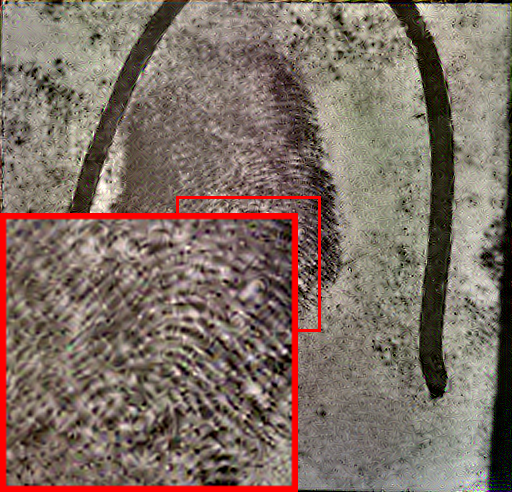}}
\subfigure[VGG conv5-3]{\label{fig:VisualRes_c}\includegraphics[width=0.4\linewidth]{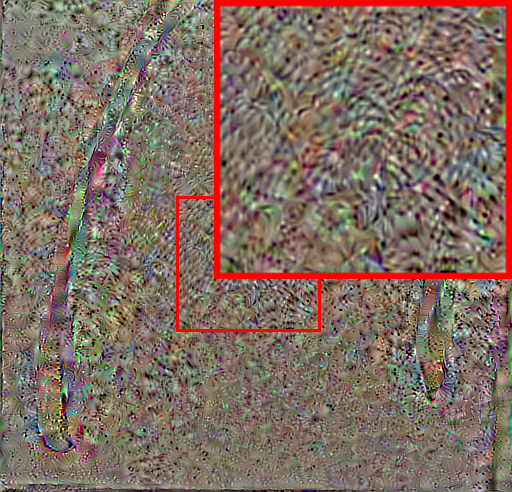}}
\subfigure[trained conv5-3]
{\label{fig:VisualRes_d}\includegraphics[width=0.4\linewidth]{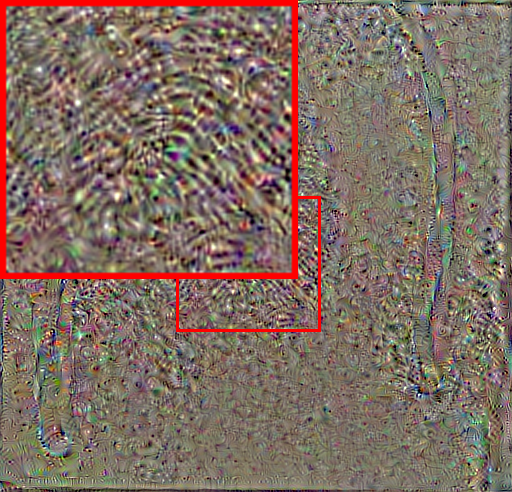}}
\caption{Comparison of visualization results. Red box shows the detailed part of the full image. (a) shows the original latent fingerprint. (b) (c) (d) are the reconstruction results from different networks and layers. }
\label{fig:VisualRes}
\end{figure}

In our approach, we take shared convolutional layers as a feature extractor, and use VGG model as a  pretrained model. To see whether we have extracted useful features, we apply visualization algorithms.

Inspired by Leon et al.~\cite{gatys2015neural}, we visualize the information by directly reconstructing the image from certain activations.

For any input image $X$ and a certain group of network activations $Y$. Our target is to find a reconstructed image $\hat{X}$ with its corresponding activations $\hat Y$ and satisfied:
\begin{equation}
\min_{\hat{X}} Loss(Y, \hat{Y})
\label{eq:Visual}
\end{equation}
In the experiment, $\hat{X}$ is zero initiated and Euclidean distance is taken as the loss function. We use L-BFGS~\cite{nocedal2006numerical} to optimize Eq.~\ref{eq:Visual} and take the result after fixed 1000 iterations.

As $\hat{X}$ and $\hat{Y}$ are zero vectors at first, large activations in $Y$ contributes more in the Euclidean distance loss function and instinctively converge earlier.

Experiment results are shown in Fig.~\ref{fig:VisualRes}. Fig.~\ref{fig:VisualRes_a} is a random picked latent fingerprint in Nist27 database. Fig.~\ref{fig:VisualRes_b} shows that VGG has the ability to reconstruct fingerprints and it's complex texture using activations in layer conv4-3. However in layer conv5-3, which we used as the last shared convolutional layers, Fig.~\ref{fig:VisualRes_c} shows that VGG failed to reconstruct fingerprint. After training, we can see a significant improvement shown in Fig.~\ref{fig:VisualRes_d}. So that our net successfully extracted ridges and valleys of a latent fingerprint.

Summing up, although VGG is trained on nature images, it shows the ability to represent fingerprints. After finetuning, our net learns to extract ridge and valley features in a latent fingerprints.

\section{Conclusion and future work}
We propose a novel algorithm for efficient and reliable minutiae extraction on latent fingerprints.
Fully convolutional network learns to generate minutiae location proposals from raw latent fingerprint directly. Then a CNN shared convolutional layers with the FCN above is utilized to classify proposals and calculate minutia orientations. Minutia descriptor is learned end-to-end and the whole process is fast.

Future work will include (1) using more powerful networks, (2) adding image enhancement to preprocess latent fingerprints, and (3) matching fingerprints on feature maps.

\section*{Acknowledgment}
This work was supported by NSFC(61333015).


\begin{thebibliography}{10}\itemsep=-1pt

\bibitem{cao2015latent}
K.~Cao and A.~K. Jain.
\newblock Latent orientation field estimation via convolutional neural network.
\newblock In {\em Biometrics (ICB), 2015 International Conference on}, pages
  349--356. IEEE, 2015.

\bibitem{cao2014segmentation}
K.~Cao, E.~Liu, and A.~K. Jain.
\newblock Segmentation and enhancement of latent fingerprints: A coarse to fine
  ridgestructure dictionary.
\newblock {\em Pattern Analysis and Machine Intelligence, IEEE Transactions
  on}, 36(9):1847--1859, 2014.

\bibitem{feng2014high}
J.~Feng, C.~Liu, H.~Wang, and B.~Sun.
\newblock High-resolution palmprint minutiae extraction based on gabor feature.
\newblock {\em Science China Information Sciences}, 57(11):1--15, 2014.

\bibitem{garris2000nist}
M.~D. Garris and R.~M. McCabe.
\newblock Nist special database 27: Fingerprint minutiae from latent and
  matching tenprint images.
\newblock {\em National Institute of Standards and Technology, Technical Report
  NISTIR}, 6534, 2000.

\bibitem{gatys2015neural}
L.~A. Gatys, A.~S. Ecker, and M.~Bethge.
\newblock A neural algorithm of artistic style.
\newblock {\em arXiv preprint arXiv:1508.06576}, 2015.

\bibitem{girshick2015fast}
R.~Girshick.
\newblock Fast r-cnn.
\newblock In {\em Proceedings of the IEEE International Conference on Computer
  Vision}, pages 1440--1448, 2015.

\bibitem{girshick2014rich}
R.~Girshick, J.~Donahue, T.~Darrell, and J.~Malik.
\newblock Rich feature hierarchies for accurate object detection and semantic
  segmentation.
\newblock In {\em Proceedings of the IEEE conference on computer vision and
  pattern recognition}, pages 580--587, 2014.

\bibitem{he2015deep}
K.~He, X.~Zhang, S.~Ren, and J.~Sun.
\newblock Deep residual learning for image recognition.
\newblock {\em arXiv preprint arXiv:1512.03385}, 2015.

\bibitem{jain1997line}
A.~Jain, L.~Hong, and R.~Bolle.
\newblock On-line fingerprint verification.
\newblock {\em Pattern Analysis and Machine Intelligence, IEEE Transactions
  on}, 19(4):302--314, 1997.

\bibitem{jia2014caffe}
Y.~Jia, E.~Shelhamer, J.~Donahue, S.~Karayev, J.~Long, R.~Girshick,
  S.~Guadarrama, and T.~Darrell.
\newblock Caffe: Convolutional architecture for fast feature embedding.
\newblock {\em arXiv preprint arXiv:1408.5093}, 2014.

\bibitem{long2015fully}
J.~Long, E.~Shelhamer, and T.~Darrell.
\newblock Fully convolutional networks for semantic segmentation.
\newblock In {\em Proceedings of the IEEE Conference on Computer Vision and
  Pattern Recognition}, pages 3431--3440, 2015.

\bibitem{nocedal2006numerical}
J.~Nocedal and S.~Wright.
\newblock {\em Numerical optimization}.
\newblock Springer Science \& Business Media, 2006.

\bibitem{ren2015faster}
S.~Ren, K.~He, R.~Girshick, and J.~Sun.
\newblock Faster r-cnn: Towards real-time object detection with region proposal
  networks.
\newblock In {\em Advances in Neural Information Processing Systems}, pages
  91--99, 2015.

\bibitem{rijsbergencj}
V.~Rijsbergen.
\newblock Cj information retrieval. 1979.

\bibitem{sankaran2014latent}
A.~Sankaran, P.~Pandey, M.~Vatsa, and R.~Singh.
\newblock On latent fingerprint minutiae extraction using stacked denoising
  sparse autoencoders.
\newblock In {\em Biometrics (IJCB), 2014 IEEE International Joint Conference
  on}, pages 1--7. IEEE, 2014.

\bibitem{sermanet2013overfeat}
P.~Sermanet, D.~Eigen, X.~Zhang, M.~Mathieu, R.~Fergus, and Y.~LeCun.
\newblock Overfeat: Integrated recognition, localization and detection using
  convolutional networks.
\newblock {\em arXiv preprint arXiv:1312.6229}, 2013.

\bibitem{simonyan2014very}
K.~Simonyan and A.~Zisserman.
\newblock Very deep convolutional networks for large-scale image recognition.
\newblock {\em arXiv preprint arXiv:1409.1556}, 2014.

\bibitem{uijlings2013selective}
J.~R. Uijlings, K.~E. van~de Sande, T.~Gevers, and A.~W. Smeulders.
\newblock Selective search for object recognition.
\newblock {\em International journal of computer vision}, 104(2):154--171,
  2013.

\bibitem{watson1992nist}
C.~Watson and C.~Wilson.
\newblock Nist special database 4.
\newblock {\em Fingerprint Database, National Institute of Standards and
  Technology}, 17:77, 1992.

\bibitem{watson2007user}
C.~I. Watson, M.~D. Garris, E.~Tabassi, C.~L. Wilson, R.~M. Mccabe, S.~Janet,
  and K.~Ko.
\newblock User's guide to nist biometric image software (nbis).
\newblock 2007.

\bibitem{yang2014localized}
X.~Yang, J.~Feng, and J.~Zhou.
\newblock Localized dictionaries based orientation field estimation for latent
  fingerprints.
\newblock {\em Pattern Analysis and Machine Intelligence, IEEE Transactions
  on}, 36(5):955--969, 2014.

\bibitem{zeiler2014visualizing}
M.~D. Zeiler and R.~Fergus.
\newblock Visualizing and understanding convolutional networks.
\newblock In {\em Computer vision--ECCV 2014}, pages 818--833. Springer, 2014.

\end{thebibliography}
\end{document}